\documentclass[11pt]{article}

\usepackage[preprint]{acl}

\usepackage{times}
\usepackage{latexsym}
\usepackage{mathptmx}
\usepackage{float}
\usepackage{todonotes}
\usepackage{outlines}
\usepackage{algpseudocode}
\usepackage{booktabs}
\usepackage{subcaption}

\usepackage[T1]{fontenc}

\usepackage[utf8]{inputenc}

\usepackage{microtype}

\usepackage{inconsolata}

\usepackage{graphicx}

\usepackage{enumitem}
\setitemize{noitemsep,topsep=0pt,parsep=0pt,partopsep=0pt}

\title{Peek2: Regex-free Byte-level Byte-Pair Encoding Pretokenizer for LLM Inference on Edge Devices}

\author{Liu Zai \and Iraklis Klampanos \\
        University of Glasgow \\ University Avenue \\ Glasgow G12 8QQ}

\begin{document}
\maketitle
\begin{abstract}
Pretokenization is a crucial, sequential pass in Byte-level BPE tokenizers, yet little work has been done to optimize it for edge-side inference. Our proposed new implementation, Peek2, serves as a drop-in replacement for cl100k-like pretokenizers used in GPT-3, LLaMa-3, and Qwen-2.5. After breaking down and analyzing the logic of the original cl100k pretokenizer, we introduced a new pretokenization algorithm with linear time complexity and constant, trivial memory usage, suited for edge scenarios. Test results show that it increases microbenchmarking throughput by up to \( 2.48\times \) and delivers a \( 1.14\times \) improvement in overall throughput across the entire Byte-level BPE encoding process, depending on the dataset, while providing identical results as the baseline Regex-based tokenizer.
\end{abstract}

\section{Introduction}

Byte-level Byte-Pair Encoding (BPE) tokenizers \cite{sennrich_2016_neural} are widely used in Large Language Models (LLMs) to transform raw text into a sequence of tokens. A pretokenizer is used at the start of the tokenization process, segmenting the original text into shorter fragments to perform pair-merging individually. In practice, the pretokenizer is often implemented using Regular Expressions (Regex).

Recent advances in edge–cloud hybrid LLM inference systems have begun to challenge the long-standing dominance of server-centric architectures \cite{hybrid-1} \cite{hybrid-2}. Improvements in model compression and runtimes now enable portions of LLM workloads to run directly on edge platforms such as desktop PCs, laptops, and embedded devices. We found that these systems perform BPE on low-cost edge devices, including pretokenization. For server-side training and inference, compiling and executing complex Regex is acceptable. However, on edge platforms, the Regex-based pretokenizer may introduce additional overhead that can affect throughput and cold-start time, due to low processing power, limited memory, and the lack of certain instruction sets.

As presented by \citet{hybrid-1}, the model workload can be routed to the cloud or SLM on the edge after tokenization. With prior work such as BlockBPE \cite{blockbpe}, the pair-merging process is also parallelizable, though porting it to edge systems remains necessary. From an overall perspective, the pretokenizing phase, which happens before the pair-merging process, becomes the final piece left sequential and unoptimized. This motivates us to seek an optimization for the widely adopted pretokenizer, cl100k \cite{gpt3}.

While optimizing the inference pretokenizer, it is necessary to maintain bug-for-bug compatibility with the training pretokenizer. The segment boundaries inserted during the pretokenize phase are carried over to the pair-merging phase. If the pretokenizer is simplified without retraining, a disparity between the inference and training systems would arise, reduce the overall BPE compression rate, and negatively impact downstream performance \cite{schmidt-etal-2024-tokenization}.

\begin{figure}[htb]
    \centering
    \begin{subfigure}{0.4\linewidth}
        \includegraphics[width=\linewidth]{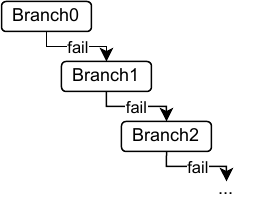}
        \caption{Regex}
        \label{fig:cl100k_design_original}
    \end{subfigure}
    \begin{subfigure}{0.4\linewidth}
        \includegraphics[width=\linewidth]{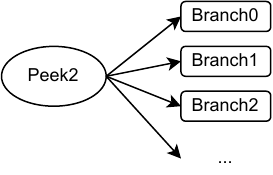}
        \caption{Peek2}
        \label{fig:cl100k_design_peek2}
    \end{subfigure}
    \caption{Process diagram of different implementations of Cl100k}
    \label{fig:cl100k_designs}
\end{figure}

Existing pretokenizer optimizations are usually server-oriented \cite{blockbpe}. Aiming to improve the overall throughput of the tokenization process at the edge, we propose a reimplementation of the cl100k \cite{tiktoken} pretokenizer, which we refer to as Peek2.

\section{Background: Cl100k Analysis}

Cl100k \cite{tiktoken} is the pretokenizer introduced with GPT-3 \cite{gpt3}. It is then widely adopted by other LLM tokenizers. LLaMa-3 \cite{llama3} and Qwen-2.5 \cite{qwen2.5} are examples of such models. The pretokenizer segments the script by repeatedly applying the Cl100k regex pattern, shown in Appendix~\ref{sec:cl100k_regex}, inserting a break after each match.

The pretokenizer Regex is formed in an or-of-branches hierarchy. The text is greedily matched against the leftmost branch, and if the match fails, the next branch is tested, as shown in Figure~\ref{fig:cl100k_design_original}.

\subsection{Left Snapping}

For convenience, we introduce \texttt{Left Snapping}. In this paper, the \verb~|~ symbol is used to symbolize a break inserted in text, e.g.

\verb~Lorem| ipsum| dolor| sit| amet|.~

Splits the original script into \( 6 \) individual parts. Note that the space is combined with the next word to form a segment.

\subsection{Functionality of Each Branch}

\paragraph{Branch0 handles contractions.} It matches common contractions in common Latin scripts.

\paragraph{Branch1 handles words.} Matches any group of letters, usually forming a word in Latin scripts, or a subsentence in East Asian scripts. It \texttt{Left Snaps} one character not of Unicode Letter Class, Unicode Number Class, ASCII CR, or ASCII LF.

\paragraph{Branch2 handles numbers.} The numbers are grouped into sets of $3$ digits.

\paragraph{Branch3 handles punctuations and other characters.} Anything that does not match the Unicode Letter Class, Unicode Number Class, or ASCII Whitespaces is grouped together. This cluster will also \texttt{Left Snap} one space, and \texttt{Right Snap} any count of line folds.

\paragraph{Branch4 handles whitespaces.} Whitespaces are grouped together, with the following exceptions: Will break after the last one of a cluster of line folds. Will break before the last whitespace, enabling the next word to left snap that whitespace if it is there.

\section{Our proposed pretokenizer: Peek2}

We present an optimized pretokenizer implementation, Peek2, primarily focused on improving the branch decision process, as shown in Figure~\ref{fig:cl100k_design_peek2}.

\subsection{Categorizing the Peeked Character}

In the original Regex, there are multiple match elements: some detect the Unicode scalar value, while others classify whether it belongs to a predefined Unicode Class. By applying a process similar to alphabet compression, we get an exhaustive list of categories, which provides just-enough granularity for pretokenization:

\begin{outline}
  \1 Category0: All other scalars
  \1 Category1: ASCII Space
  \1 Category2: ASCII Single Quote
  \1 Category3: ASCII CR or LF
  \1 Category4: Unicode Letter Class
  \1 Category5: Unicode Whitespace Class
  \1 Category6: Unicode Number Class
\end{outline}

To reduce duplicated lookups, we define PeekCategorize in Figure~\ref{alg:peekcategorize}, which takes a Unicode scalar value and returns its category. The smaller subsets take precedence, i.e., Category1 will be returned instead of Category5 for the ASCII space \verb|' '| character, and a match group targeting the Unicode Whitespace Class should accept Category1, Category3, and Category5.

\begin{figure}[htb]
\small
\begin{algorithmic}
\Procedure{PeekCategorize}{$scalar$}
    \If{$scalar = Space$}
        \Return 1
    \ElsIf{$scalar = SingleQuote$}
        \Return 2
    \ElsIf{$scalar = Cr | Lf$}
        \Return 3
    \Else
        \State $category \gets \Call{UnicodeClassOf}{scalar}$
        \If{$category = Letter$}
            \Return 4
        \ElsIf{$category = Whitespace$}
            \Return 5
        \ElsIf{$category = Number$}
            \Return 6
        \EndIf
    \EndIf
    \State \Return 0
\EndProcedure
\end{algorithmic}
\caption{PeekCategorize, classifying the peeked Unicode scalar value}
\label{alg:peekcategorize}
\end{figure}

\subsection{Branch Decision}

Because only one character can be snapped at a time, the second character's category instantly determines the logic path: it either triggers a \texttt{Snap} for the first character or acts as a continuation of the first character's segment.

Based on PeekCategorize (Figure~\ref{alg:peekcategorize}), the two branch-deciding temporal steps now have \( 7 \times 7 \) input states. This makes it applicable for replacement with a one-time table lookup, as opposed to the original Unicode Scalar input space, which is prohibitively large (\( 1e5 ^ 2 \) level). This replacement is inspired by Hopcroft's DFA Minimization Algorithm \cite{partition-refinement}, which iteratively splits state partitions until no distinguishable states remain. Other temporal steps can be replaced by specific branch functions that trivially look for the next segment stop without any failover. 

We have already labelled branches in the original Regex implementation with unique integer indices, from Branch0 to Branch6. For each branch, we translate the first two match sets into the corresponding list of PeekCategorize-returned binary tuples of integers. Each character pair (tuple) is mapped to a specific rule ID (branch index) in Table~\ref{tab:branch_table}. To maintain the original Regex priority, cells are filled in order; once a high-priority rule (a smaller index) claims a cell, it cannot be overwritten by a later one.

\begin{table}[htb]
\small
\centering
\begin{tabular}{lrrrrrrr}
\toprule
          &  \multicolumn{7}{r}{Cat1} \\
Cat0      & 0 & 1 & 2 & 3 & 4 & 5 & 6 \\
\midrule
0         & 3 & 3 & 3 & 3 & 1 & 3 & 3 \\
1         & 3 & 4 & 3 & 4 & 1 & 4 & 4 \\
2         & 3 & 3 & 3 & 3 & 0 & 3 & 3 \\
3         & 4 & 4 & 4 & 4 & 4 & 4 & 4 \\
4         & 1 & 1 & 1 & 1 & 1 & 1 & 1 \\
5         & 4 & 4 & 4 & 4 & 1 & 4 & 4 \\
6         & 2 & 2 & 2 & 2 & 2 & 2 & 2 \\
\bottomrule
\end{tabular}
\caption{Branch Decision Lookup Table}
\label{tab:branch_table}
\end{table}

The result in Table~\ref{tab:branch_table} maps identically to the previously discussed branch-fallback logic. Pretokenization is achieved in Figure~\ref{alg:pretokenize} by recursively peeking at the next two scalars and using this table to call the appropriate function to segregate and remove the next segment from the remaining string.

\begin{figure}[htb]
\small
\begin{algorithmic}
\Procedure{Pretokenize}{$string$}
    \State $Cat0 \gets \Call{PeekCategorize}{string[0]}$
    \State $Cat1 \gets \Call{PeekCategorize}{string[1]}$
    \State $string \gets \Call{LookupAndCallBranch}{Cat0, Cat1, string}$
    \State $\Call{Pretokenize}{string}$
\EndProcedure
\end{algorithmic}
\caption{Pretokenize, the core pretokenizer algorithm}
\label{alg:pretokenize}
\end{figure}

\subsection{Handling of Special Cases}

There is an exception of fallbacks happening later: during the handling of Branch0, if the subsequent pattern does not match any common contractions, the single quote should fallback to Category0 and pair with the next letter to fallback to Branch1. We can simply let Branch0 handler function chain invoke Branch1 handler function to handle this exception.

\subsection{Demonstration by Example}

Given an example input string presegmented:

\verb'Color| :| Red'

Now we demonstrate and compare an Non-deterministic Finite Automaton (NFA)-based Regex engine and our Peek2 algorithm for cutting out the secondary segment \verb|" :"|.

The Regex engine will match Branch0 to Branch6 one by one. While other branches fail at the first character, Branch1 and Branch3 share the \texttt{Left Snapping} behavior, so after identifying the first space, the next scalar is first tested if it is of the Unicode Letter class as required by Branch1, which it is not. A stack-like structure is used for failback, and the two characters will be tested again when Branch3 is reached.

Our proposed Peek2 algorithm will first run PeekCategorize on the space and the colon simultaneously. The space is categorized as Category1, and the colon will be Category0. After looking up Table~\ref{tab:branch_table}, the correct Branch3 is selected.

After the branch decision, the branch function trivially replicates the remaining matching logic performed by the Regex engine until the next mismatch, then the branch selection comes again. Since the branch decision is always identical, the bug-for-bug compatibility is guaranteed.

\subsection{Time and Space Complexity}

As in the previous example, normally each character is visited exactly once in branch decisions. On rare occasions, such as the late fallback situation described earlier, a small portion of characters might be visited \( 2 \) times. So, the time complexity of Peek2 is \( O(n) \), where \( n \) is the length of the input sequence. The space complexity of Peek2 is strictly \( O(1) \) as the lookup table has a fixed size.

For comparison, as the Regex expression gets complex with multiple match groups, it is found \cite{bille_regex_complexity} that the time complexity of NFA-based Regex engines \cite{ken_regex} is \( O(n \times k) \) or even \( O(n \times m) \), where \( k \) is the number of groups and \( m \) is the length of the pattern. Deterministic Finite Automaton (DFA)-based Regex engines can achieve linear time complexity \cite{cox_regex}, but require a complex compiling stage, which often results in high memory usage.

\section{Test Results}

We implemented the Peek2 pretokenizer with Safe Rust, then integrated this approach into the open-source Hugging Face tokenizers \cite{hf_tokenizers} library on top of version \verb|0.22.3-dev.0|. The tests are carried out over four datasets: \verb|en|, \verb|cn|, \verb|code|, and \verb|math|, respectively, collected from The FineWeb Datasets \cite{fineweb}, Fineweb-Edu-Chinese \cite{fineweb_edu_chinese}, codesearchnet \cite{codesearchnet}, and opc-fineweb-math \cite{opc-fineweb-math}.

All of our test cases involve a $10$-second warm-up period. The target task is then run over the target dataset repeatedly for $30$ seconds, collecting the duration for each sample and the overall throughput. The tests are run on an Intel Core i5-13600KF CPU.

\subsection{Microbenchmarking}

First, we perform a microbenchmark of the pretokenization stage. No downstream tasks are included in the microbenchmarking.

\begin{figure}[htb]
    \includegraphics[width=\linewidth]{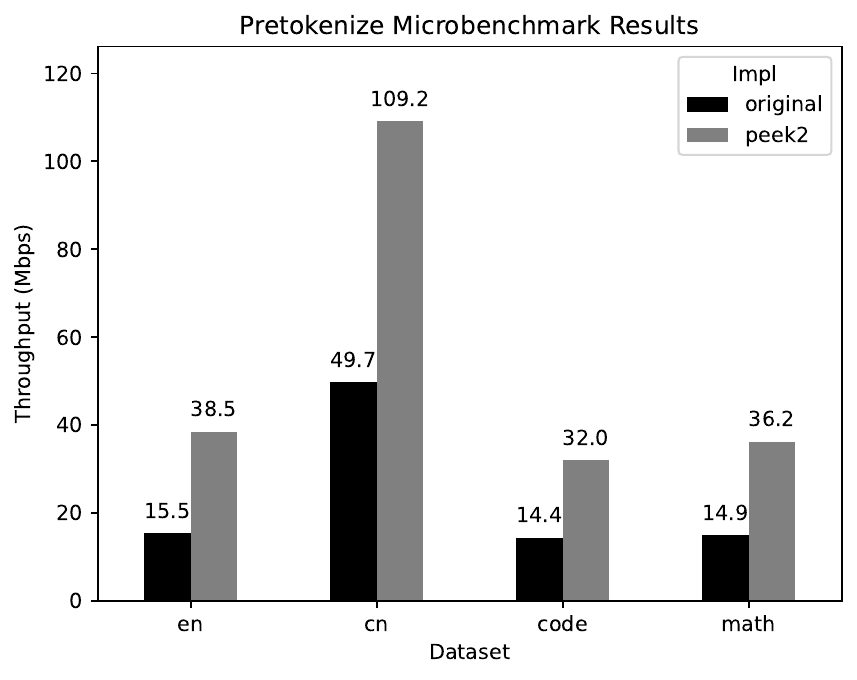}
    \caption{Throughput comparison of the pre-tokenization stage (errors $ <1\% $) }
    \label{fig:microbench}
\end{figure}

The throughput has increased sharply, as shown in Figure~\ref{fig:microbench}. All datasets have more than doubled their throughput, with the \verb|en| dataset achieving the largest gain, up to \( 2.48\times \). The \verb|cn| dataset shows the least gain (\( 2.20\times \)), as East Asian scripts have only sub-sentence splitting during this stage, unlike other datasets, which undergo word-level tokenization.

Across all four datasets, the pretokenizer yields the same results as the Regex-based splitter. This validates the bug-for-bug compatibility we aimed for in design.

\subsection{End-to-End Benchmarking}

Next, we tested and benchmarked the Peek2 pretokenizer versus original Regex pretokenizer across a range of tasks of the complete LLaMa-3 BPE pipeline. The datasets are split into batches of \( 1000 \) and distributed to multiple threads to simulate real-world scenarios of long context handling and BPE training.

\begin{figure}[htb]
    \includegraphics[width=\linewidth]{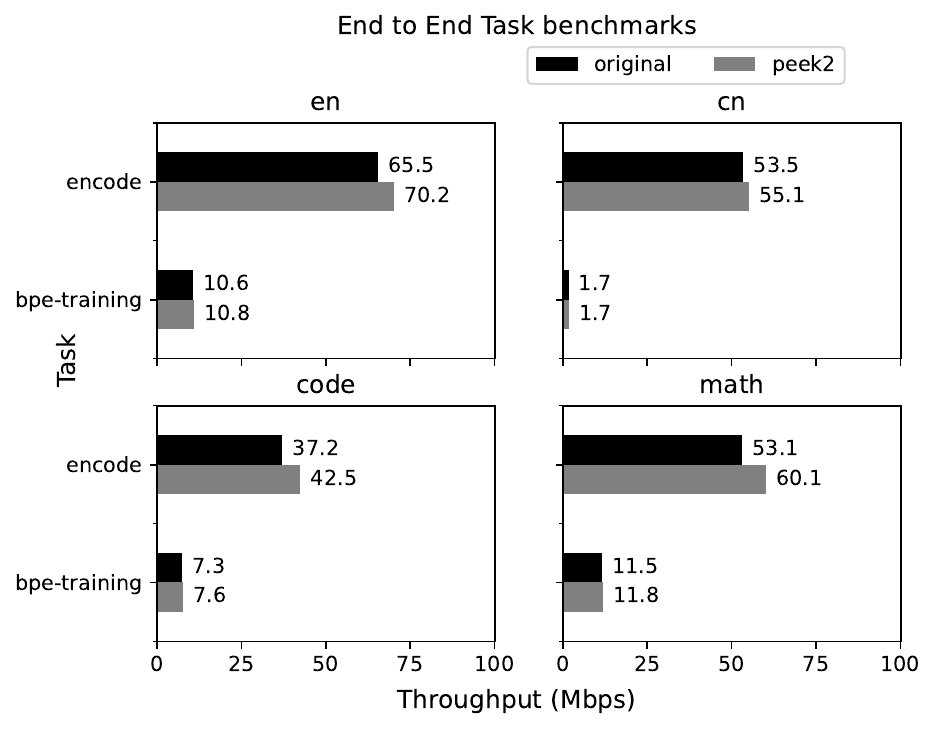}
    \caption{Throughput comparison of different end-to-end tasks (errors $ <1\% $) }
    \label{fig:e2e}
\end{figure}

All encoding tasks improved their throughput by \( 1.03\times \) to \( 1.14\times \), as shown in Figure~\ref{fig:e2e}. However, across all BPE training tasks, gains are much smaller than those in encoding, as downstream tasks dominate the process. \verb|code| and \verb|math| have the most significant improvements in their encoding tasks. This is expected, as code and math-heavy data rely more on pretokenization, as there are fewer subword units requiring downstream pair merging.

\begin{table}[htb]
\small
\centering
\begin{tabular}{lrr}
\toprule
 & \multicolumn{2}{r}{Pretokenize Time (\%)} \\
Impl & original & peek2 \\
Dataset &  &  \\
\midrule
en & 11.1 & 6.9 \\
cn & 5.4 & 3.0 \\
code & 22.8 & 14.0 \\
math & 19.8 & 12.7 \\
\bottomrule
\end{tabular}
\caption{Pretokenization time consumption proportion, for encoding tasks, comparing the original pretokenization with Peek2}
\label{tab:pretok_percent}
\end{table}

Because the \verb|encode| task process shares the same input with microbenchmarks, we calculate the time proportion of the pretokenization process relative to the full end-to-end pipeline, as presented in Table~\ref{tab:pretok_percent}. While the \verb|cn| dataset is insensitive to pretokenizer optimizations, in other datasets, especially \verb|code|, pretokenizer performance plays a significant role.

\section{Conclusion}

We present Peek2, an edge-optimized, Regex-free implementation of the cl100k pretokenizer, designed around a branch decision table that replaces the branch failback logic in Regex with a one-time lookup.

Test results show that while maintaining identical output, higher throughput is achieved on all tokenizer-related tasks. This work can serve as a drop-in replacement for GPT-3 \cite{gpt3}, LLaMa-3 \cite{llama3}, and Qwen-2.5 \cite{qwen2.5} pretokenizers for edge or edge-cloud hybrid LLM inference systems, for better tokenization throughput.

\section*{Limitations}

Peek2 is solely based on the currently widely adopted cl100k-like pretokenizers. Future research might migrate away to other pretokenizers for better performance of the BPE and the LLM. However, this optimization paradigm might be reused for future research.

Currently, the experiments are limited to desktop CPUs. Although we expect the results to be similar, if not better, future work could add comparisons of laptops and embedded devices to cover more platforms for edge inference.

Peek2 is a CPU algorithm. Future work could port it to TPUs and APUs, if available on edge processors, to improve throughput and utilization.

Peek2 was designed to be bug-for-bug compliant with the Regex implementation. However, some of the bugs are almost certainly introduced by Regex's ambiguity. Notably:

\verb~'D|oes| it| work|?’| She| asked.~

Notice how the \verb~'D~ is split out due to misinterpretation as a contraction. Evaluation of fixing such bugs might involve retraining or post-training the BPE, or even the model itself, which is out of the scope of this paper.

\bibliography{custom}

\clearpage

\section*{Appendix}

\appendix

\section{Cl100k Regex}
\label{sec:cl100k_regex}

\subsection{Complete Regex}

\begin{verbatim}
'(?i:[sdmt]|ll|ve|re)|
[^\r\n\p{L}\p{N}]?+\p{L}++|
\p{N}{1,3}+|
 ?[^\s\p{L}\p{N}]++[\r\n]*+|
\s++$|\s*[\r\n]|\s+(?!\S)|\s~
\end{verbatim}

The linefolds are added for the ease of interpretation.

\subsection{Verbose Branches}

To improve the clarity of this complex Regex, we also provide each Branch in Python's VERBOSE Regex format, with additional whitespaces added for visual guidance. The original whitespaces are escaped with \verb|\|.

\subsubsection{Branch 1}

\begin{verbatim}
  '
  (?:
      [sdmt]
      | ll
      | ve
      | re
  )
\end{verbatim}

\subsubsection{Branch 2}

\begin{verbatim}
  [^\r\n\p{L}\p{N}]?+
  \p{L}++
\end{verbatim}

\subsubsection{Branch 3}

\begin{verbatim}
  \p{N}{1,3}+
\end{verbatim}

\subsubsection{Branch 4}

\begin{verbatim}
  \ ?
  [^\s\p{L}\p{N}]++
  [\r\n]*+ 
\end{verbatim}

\subsubsection{Branch 5}

\begin{verbatim}
    \s++$
  | \s*[\r\n]
  | \s+(?!\S)
  | \s~
\end{verbatim}

\section{Code and Data Availability}

\subsection{GitHub Link}

\url{https://github.com/omegacoleman/tokenizers_peek2}

\subsection{Archival DOI}

\url{https://doi.org/10.5281/zenodo.18459917}

\end{document}